\documentclass[runningheads]{llncs}
\usepackage{amsfonts}       
\usepackage{amsmath}
\usepackage{graphicx}
\usepackage{caption}
\usepackage{subcaption}
\usepackage{multirow}

\usepackage[linesnumbered,ruled,vlined]{algorithm2e}
\SetKwInput{KwInput}{Input}                
\SetKwInput{KwOutput}{Output} 

\begin{document}
%
\title{Distributed Interaction Graph Construction for Dynamic DCOPs in Cooperative Multi-agent Systems}
%
%
\author{Brighter Agyemang \and
Fenghui Ren \and
Jun Yan}

\authorrunning{B. Agyemang et al.}
%
%

\institute{School of Computing and Information Technology, University of Wollongong, Wollongong, Australia \\\email{ba233@uowmail.edu.au, \{fren, jyan\}@uow.edu.au}
}

\maketitle              
\begin{abstract}

DCOP algorithms usually rely on interaction graphs to operate. In open and dynamic environments, such methods need to address how this interaction graph is generated and maintained among agents. Existing methods require reconstructing the entire graph upon detecting changes in the environment or assuming that new agents know potential neighbors to facilitate connection. We propose a novel distributed interaction graph construction algorithm to address this problem. The proposed method does not assume a predefined constraint graph and stabilizes after disruptive changes in the environment. We evaluate our approach by pairing it with existing DCOP algorithms to solve several generated dynamic problems. The experiment results show that the proposed algorithm effectively constructs and maintains a stable multi-agent interaction graph for open and dynamic environments.

\end{abstract}
\section{Introduction}
The complexities of computational problems in our world today make it increasingly appealing to, where possible, address such problems using collaborative autonomous agents. To this end, the domain of multi-agent collaboration has been studied from different perspectives in recent years. Some interesting application domains include power systems, mobile sensing~\cite{Yedidsion2018}, disaster management~\cite{973385}, environment monitoring~\cite{10.1145/1160633.1160885,10.1007/978-3-642-40643-0_21}, and traffic light management~\cite{10.5555/1402298.1402308}.

In several of these application domains, the environment is dynamic~\cite{Nguyen2012}, and the interactions between the agents evolve over the problem horizon. Thus, predefined interaction structures may not be useful. It is then significant to address how the agents interact in a dynamic environment to facilitate the application of algorithms that leverage such agent structures. In~\cite{Sultanik2009}, this is referred to as the Dynamic Distributed Multi-agent Hierarchy Generation ((DynDisMHG)) problem. There exist other works in the literature with similar motivations~\cite{Yeoh2015,Petcu2005,Yeoh2011,Barambones2021}. Multi-agent hierarchies enable the application of certain Distributed Constraint Optimization Problem (DCOP) proposals for multi-agent collaboration. Additionally, multi-agent hierarchies have applications in mobile sensing agents and mobile ad-hoc networks.

One common assumption among existing approaches is that there is an expected interaction graph to enable the generation of the multi-agent hierarchy. In open and dynamic environments such as a multi-agent rescue system or a Mobile Sensor Team (MST), an agent's interactions can vary over time, and it may be introduced or removed from the environment at arbitrary periods~\cite{Yedidsion2018}. Therefore, the agents will have to be equipped to generate and maintain the multi-agent hierarchy collaboratively.

In this study, we discuss the Ad-hoc Distributed Multi-agent Hierarchy generation problem as an extension of the (DynDisMHG) problem and propose an algorithm for this class of problems in Multi-Agent Systems (MASs). We focus on applying our approach in the DCOP domain and show the proposed method's effectiveness via experiments. The proposed method has application in areas such as DCOP\_MSTs~\cite{Zivan2015,Yedidsion2018} where agent positions or environment may change and agent interactions have to be defined to enable the use of DCOP methods.

In what follows, we discuss related work in Section~\ref{sec:related-work}. Section~\ref{sec:problem-formulation} introduces the background and problem formulation of this study. Section~\ref{sec:proposed-algorithm} discusses our proposed approach. In Section~\ref{sec:experiments}, we first introduce the experiment setup and then discuss the results. We draw our conclusions in Section~\ref{sec:conclusion}.

\section{Related Work}\label{sec:related-work}

A well-known study in D-DCOP is the work in~\cite{YoussefHamadiChristianBessiere1998} where a Distributed DFS (DDFS) algorithm was proposed for ordering variables (or agents) in a Distributed Constraint Satisfaction Problem (DCSP). For each agent, DDFS classifies a known neighbor list, retrieved from a predefined constraint graph, into parent and children. In an open and dynamic environment where the constraint graph changes, DDFS recomputes the interaction graph from scratch~\cite{Yeoh2015}.

The DDFS algorithm has been used in many notable D-DCOP algorithms. The work in~\cite{Yeoh2015} proposed HARP for detecting agents affected by constraint changes and then invokes DDFS to reconstruct the hierarchy. Other DFS proposals exist~\cite{COLLIN1994297} and have been used with algorithms in the DCOP domain~\cite{Petcu2005,Petcu2007}. DDFS algorithms generate the hierarchy from scratch with every change, which unnecessarily causes unaffected parts of the hierarchy to be reconstructed.

Also,~\cite{Sultanik2009} proposed Mobed for maintaining multi-agent hierarchies. The proposed algorithm ensures that an insertion point is unique for a joining agent and that the resulting tree is valid, a concept we adopt in our proposal. Mobed requires an expected interaction graph and an initial multi-agent hierarchy.

In summary, DDFS algorithms require a restart upon a change to generate valid multi-agent hierarchies, whereas methods like Mobed and HARP are hierarchy maintenance methods. The restart is necessary since the ordering heuristic may be inconsistent with the hierarchy after a change, and there is no procedure for introducing a new agent. Also, the methods mentioned above assume the existence of an interaction graph applicable across the problem horizon. On the other hand, we consider an ad-hoc multi-agent hierarchy generation problem and propose a valid hierarchy generation and maintenance method.

\section{Problem Formulation And Background}\label{sec:problem-formulation}

\subsection{Problem Formulation}\label{subsec:adhoc-def}
We extend the formulation in~\cite{Sultanik2009} to the domain where no fixed interaction graph is guaranteed to be consistent across the horizon of the MAS. Let $A_t$ be an ordered set of agents in a multi-agent system at time $t$ that gives rise to an unordered, labeled, rooted tree. We call the tree $T=\left<A, \pi:A\rightarrow A\right>$ a multi-agent hierarchy. $\pi$ is a function that, given an agent $a_j\in A$, specifies the parent of $a_j$ already in the hierarchy. The neighbor set of $a_j$ is denoted as $N_j$. We denote the children set of $a_j$ as $C_j \subseteq N_j$. The hierarchy is valid if, after adding or removing agent $a_i$ to the hierarchy, it continues to be acyclic after a finite number of steps. Such a hierarchy enables agents in disjoint parts of the tree to execute in parallel.

\subsection{Dynamic DCOP}\label{subsec:dcop}
We consider the operation of the ad-hoc multi-agent hierarchy in a dynamic DCOP context. In multi-agent systems formulated as DCOPs, agents assign values from a domain to their decision variables to optimize certain constraint functions. It is assumed that the agents are fully cooperative and can fully observe the environment~\cite{Fioretto2018,SarkerAmit;ChoudhuryMoumita;khan2021}. Also, the environment is static and deterministic. The DCOP is modeled as a tuple $P=\left<\mathbf{A},\mathbf{X},\mathbf{D},\mathbf{F},\alpha\right>$, where:
\begin{itemize}
	\item $\mathbf{A}=\left\{a_1,a_2,...,a_m\right\}$ is a finite set of agents,
	\item $\mathbf{X}=\left\{x_1,x_2,...,x_n\right\}$ is a finite set of variables,
	\item $\mathbf{D}=\left\{D_1,D_2,...,D_n\right\}$ is a set of variable domains such that, the domain of $x_i\in \mathbf{X}$ is $D_i$,
	\item $\mathbf{F}=\left\{f_1, f_2, ..., f_K\right\}$ is a set of constraint functions defined on $\mathbf{X}$ where each $f_k\in \mathbf{F}$ is defined over a subset $\mathcal{X}_k=\left\{x^k_1,x^k_2,...,x^k_p\right\}$, with $p\leq n$, determines the cost of value assignments of the variables in $\mathcal{X}_k$ as $f_k: D_1\times D_2\times ...\times D_p\rightarrow \mathbb{R}\cup \{\perp\}$. Here, the cardinality of $\mathcal{X}_k$ is the arity of $f_k$. The total cost of the values assigned to variables in $\mathbf{X}$ is $\mathbf{F}_g(\mathbf{X})=\sum_{k=1}^Kf_k(\mathcal{X}_k)$,
	\item $\alpha: \mathbf{X} \rightarrow \mathbf{A}$ is an onto function that assigns the control of a variable $x\in\mathbf{X}$ to an agent $\alpha(x)$.
\end{itemize}

We assume that $\alpha$ assigns only one variable per agent\footnote{Each variable could be represented as an aggregation of sub-variables where its domain constitutes the cartesian product of the domains of all the sub-variables. 
We use agent and variable interchangeably since an agent controls only one variable.} and the use of binary constraint functions. A Current Partial Assignment (CPA) or partial assignment is the assignment of values to a set of variables $\overline{x}$ such that $\overline{x} \subset \mathbf{X}$. A complete assignment $\sigma$ is when all variables in $\mathbf{X}$ are assigned a value. A constraint function $f_k\in\mathbf{F}$ is satisfied if $f_k(\sigma_{\mathcal{X}_k}) \neq \perp$. The objective of a DCOP is to find a complete assignment that minimizes the total cost:

\begin{equation}
	\begin{aligned}
		\sigma^* := \underset{\sigma\in\mathbf{\Sigma}}{argmin}~\mathbf{F}_g(\sigma) = \underset{\sigma\in\mathbf{\Sigma}}{argmin}~\sum_{f_k\in\mathbf{F}}f_k(\sigma_{\mathcal{X}_k}),
	\end{aligned}
	\label{eq:dcop_objective}
\end{equation}
where $\mathbf{\Sigma}$ is the set of all possible complete assignments.

The Dynamic DCOP (D-DCOP) is an extension of the DCOP formulation to address dynamic multi-agent environments. D-DCOP is modeled as a sequence of DCOPs, $\mathcal{D}_1, \mathcal{D}_2,...,\mathcal{D}_T$. Here, $\mathcal{D}_t=\left<\mathbf{A}^t,\mathbf{X}^t,\mathbf{D}^t,\mathbf{F}^t,\alpha^t\right>$ where $1\leq t \leq T$. D-DCOP aims to solve the arising DCOP problem at each time step. 

We consider adding an agent, removing an agent, and constraint function modification as events that transition the environment from one DCOP to another. Each agent $a_i$ has first to resolve its local neighbor list and parent-child associations to solve the DCOP collaboratively. 

\section{Proposed Approach}\label{sec:proposed-algorithm}
\subsection{Distributed Interaction Graph Construction Algorithm}
This study's proposed method assumes that the MAS agents can communicate via message-passing. In this message-passing approach, each message contains the sender's ID and address which the receiver can use to send a response. Other information could be contained in the message when needed. We further assume that all agents are in a cooperative team and behave cooperatively to the extent necessary for optimizing the team's utility. The agents may execute in asynchronous environments where arbitrary delay may be experienced even though message delivery is guaranteed. Therefore, we use active and inactive states to control agent procedure calls and race conditions.

The pseudocode of the proposed algorithm is presented in Algorithm~\ref{alg:digca1}. In our discussion, agent $a_i$ (referenced as $i$) is the agent that executes a procedure asynchronously. Agent $a_j$ (referenced as $j$) is an agent already in the environment that $i$ interacts with.

The agent is set to an inactive state on initialization, and other initial properties are set (line 1). The Connect function is scheduled as a process that executes regularly or based on a schedule that may depend on the application domain. When called, it first ensures that the agent is in an inactive state and that it does not have a parent (line 3). Once the connection conditions are satisfied, the agent broadcasts an Announce message in the environment (line 4) and waits for a period, condition, or timeout before proceeding (line 5). During this waiting period, an available agent in the environment receives the Announce message and responds by sending an AnnounceResponse message (lines 19-22). Before an agent can send an AnnounceResponse, it must be inactive and a response determinant function $\phi: A \rightarrow \mathbb{B}$ must be $\mathsf{TRUE}$. The function $\phi$ may be defined based on the application domain (we discuss one such function in Section~\ref{subsec:resp_det_func}).

When $i$ receives an AnnounceResponse message, it adds the sending agent to a response list if it is still in an inactive state (lines 23-26). After the waiting period (line 5), $i$ selects an agent in the response list using $\vartheta$ (line 6). This selection function, $\vartheta$, could be defined to examine each sender to determine a potential neighbor. Agent $i$ then sends an AddMe message to the selected agent $j$ and goes into an active state while it waits to hear from $j$ (lines 7-9). On line 10, it sets a timeout property to enable resetting the state on a subsequent connect call if $j$ never responds (lines 14-17), probably because it is no longer reachable. The response list is cleared before ending the execution of the connect function (line 12).

When $i$ receives an AddMe message from $j$, it adds $j$ to its children and sends a ChildAdded message to $j$ if $i$ is in an inactive state (lines 28-30). Otherwise, it replies with an AlreadyActive message to $j$ (line 33). When an AlreadyActive message is received by $i$, it sets its state to inactive (line 45) to enable the next call to the connect procedure to pass the condition on line 3.

After receiving a ChildAdded message from $j$, $i$ must be in an active state and still without a parent to proceed (line 36). If so, it assigns $j$ as its parent and goes into an inactive state (lines 37-38). Agent $i$ then sends a ParentAssigned message to $j$ (line 39). Suppose the associated DCOP algorithm's execution order is a child-first or bottom-up approach. In that case, $i$ may start the DCOP computation since its neighbor set has changed, following the notion of affected agents in~\cite{Yeoh2015} (lines 40-42).

On the other hand, when $i$ receives a ParentAssigned message, it triggers the DCOP computation if the execution order is a parent-first or top-down method (lines 46-49).

\begin{algorithm}[!htbp]
    \SetAlgoLined
    \SetKwProg{procedure}{Procedure}{}{}
    \KwInput{DCOP algorithm, stateTimeout}
    $state \gets \mathsf{INACTIVE}$, $\mathcal{L} \gets \emptyset$, $C_i \gets \emptyset$, $\pi(i) \gets null$ \tcp*[r]{initialization step}
    
    \SetKwFunction{name}{connect}
    \procedure(Called by $i$ to find a neighbor to connect to){\name{}}{
        \eIf{$state$ is $\mathsf{INACTIVE}$ and $\pi(i) = null$}{
		        Publish Announce\{$i$\} \\
		        Wait for AnnounceResponse from available agents for a period \\
		        $j \gets \vartheta(\mathcal{L})$ \tcp*{Select an agent from the list of respondents, if any}
		        \If{$j$ is found} {
		            Send AddMe\{$i$\} to $j$ \\
		            $state \gets \mathsf{ACTIVE}$ \\
		            set stateTimeout
		        }
		    $\mathcal{L} \gets \emptyset$
        } {
            \If{stateTimeout elapses and $state$ is $\mathsf{ACTIVE}$ and $\pi(i) = null$}{
                $state \gets \mathsf{INACTIVE}$ \\
                clear stateTimeout
            }
        }
        
    }
    
    \SetKwFunction{name}{receiveAnnounce} 
    \procedure(When $i$ gets an Announce message from $j$){\name{j}}{ 
        \If{$state$ is $\mathsf{INACTIVE}$ and $\phi_i(j)$} {
		  Send AnnounceResponse\{$i$,\} to $j$
	    }
    }
    
    \SetKwFunction{name}{receiveAnnounceResponse}
    \procedure(When $i$ gets AnnounceResp from $j$){\name{j}}{
        \If{$state$ is $\mathsf{INACTIVE}$} {
		  add $j$ to $\mathcal{L}$
		}
    }
    
    \SetKwFunction{name}{receiveAddMe}
    \procedure(When $i$ receives AddMe message from $j$){\name{j}}{
        \If{$state$ is $\mathsf{INACTIVE}$}{
	        add $j$ to $C_i$ \\
	        Send ChildAdded\{$i$\} to $j$
		  }
		\Else{
		    Send AlreadyActive\{$i$\} to $j$
	    }
    }
    
    \SetKwFunction{name}{receiveChildAdded}
    \procedure(When $i$ gets ChildAdded from $j$){\name{j}}{
        \If{$state$ is $\mathsf{ACTIVE}$ and $\pi(i) = null$} {
	      $\pi(i) \gets j$ \\
	       $state \gets \mathsf{INACTIVE}$ \\
	      Send ParentAssigned\{$i$\} to $j$ \\
	        \If{DCOP algorithm's execution order is $\mathsf{bottomUp}$} {
	            start DCOP algorithm
	        }
	    }
    }
    
    \SetKwFunction{name}{receiveAlreadyActive}
    \procedure(When $i$ gets AlreadyActive from $j$){\name{j}}{
        $state \gets \mathsf{INACTIVE}$
    }
    
    \SetKwFunction{name}{receiveParentAssigned}
    \procedure(When $i$ gets ParentAssigned from $j$){\name{j}}{
        \If{DCOP algorithm's execution order is $\mathsf{TopDown}$}{
		  Start DCOP algorithm
		}
    }
    
    \caption{Distributed Interaction Graph Construction Algorithm}
	\label{alg:digca1}
\end{algorithm}

In Algorithm~\ref{alg:digca1}, $i$ is able to connect to another agent in the environment or the multi-agent hierarchy described in Section~\ref{subsec:adhoc-def}. The question that must be answered is how it discovers agents in its neighbor set that are currently unreachable and update its registers accordingly. We adopt a keep-alive message approach in Algorithm~\ref{alg:digca2} to address this question.

Similar to Algorithm~\ref{alg:digca1}, the procedures in Algorithm~\ref{alg:digca2} are executed asynchronously. First, the inspectNeighbors and sendKeepAlive procedures are executed as two background processes called at regular periods like the connect procedure of Algorithm~\ref{alg:digca1}. Agent $i$ maintains a list neighbors to keep alive (line 1). The sendKeepAlive procedure sends a message to all its neighbors to inform them of its availability when called (lines 2-5). When $i$ receives a KeepAlive message, it adds the sender to the keep alive message list $P$ (lines 6-9). When the inspectNeighbors procedure is executed, it removes any neighbor $j$ that is not found in the keep-alive list (lines 11-20). If $j$ (a removed agent) was the parent, $i$ goes into an inactive state. This state change is necessary to enable the connect procedure of Algorithm~\ref{alg:digca1} to find a new parent. If a change in neighborhood is detected, $i$ starts the associated DCOP computation (lines 26-28). 

Since the execution order is a property of the DCOP algorithm, $i$ will be able to know whether to execute or forward computations to its parent or children. While we focus on DCOP computation with the multi-agent hierarchy, other distributed computations that execute using hierarchies could also be applied. Due to the asynchronous execution environment, there could be a quick succession of computation invocations. Therefore, such an environment may require computations to have an associated aborting scheme. Another approach could be to defer the execution of computations till all graph-associated message handling are complete.

\begin{algorithm}[!ht]
    \SetAlgoLined
    \SetKwProg{procedure}{Procedure}{}{}
    \KwInput{$N_i$,DCOP algorithm}
    $P \gets \emptyset$ \tcp*[r]{keep alive list}
    
    \SetKwFunction{name}{sendKeepAlive}
    \procedure(Called by $i$ to inform $j$ of its availability){\name{}}{
        \For{neighbor in $N_i$}{
		    send KeepAlive\{$i$\} to $j$
        }
    }
    
    \SetKwFunction{name}{receiveKeepAlive} 
    \procedure(When $i$ gets a KeepAlive message from $j$){\name{j}}{ 
        \If{$j$ not in $P$} {
		  Add $j$ to $P$
	    }
    }
    
    \SetKwFunction{name}{inspectNeighbors}
    \procedure(Called regularly by $i$ to remove unreachable neighbors){\name{j}}{
        $affected$ $\gets$ False \\
        \For{$j$ in $N_i$} {
		  \If{$j$ not in $P$}{
		    \If{$j$ == $\pi(i)$}{
		        $\pi(i) \gets null$ \\
		        $state \gets \mathsf{INACTIVE}$
		    }
		    \Else{
		     remove $j$ from $C_i$
		    }
		    clear all information about $j$ in $i$'s registers \\
		    affected $\gets$ True
		  }
		}
	    $P \gets \emptyset$ \\
	    \If{affected}{
	        start DCOP algorithm
	    }
    }
    
    \caption{Disconnected Neighbor Removal Algorithm}
	\label{alg:digca2}
\end{algorithm}

\begin{figure}[!t]
     \centering
     \begin{subfigure}[b]{\textwidth}
         \centering
         \includegraphics[width=.7\textwidth]{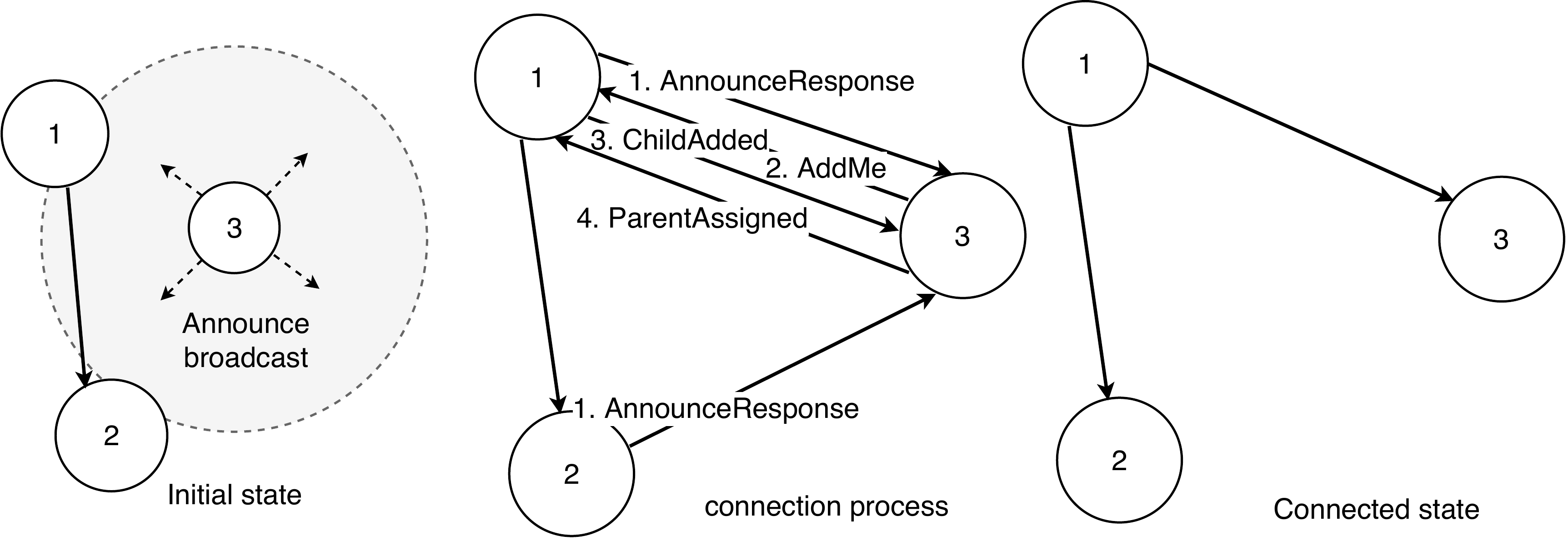}
         \label{fig:announce}
     \end{subfigure}
     
    \caption{An illustration of how an agent gets connected using the proposed algorithm. When Agent 3 wants to connect/re-connect to the current interaction graph, (1) it broadcasts Announce message to agents in range (2) Agents 1 and 2 respond to Agent 3 with AnnounceResponse, (3) Agent 3 selects agent 1 and sends an AddMe message, (4) Agent 1 responds with a ChildAdded message, (5) Agent 3 then sends a ParentAssigned message.}
    \label{fig:digca}
\end{figure}

We illustrate how an agent connects to another agent in Figure~\ref{fig:digca}. This illustration assumes a successful connection process on first try for didactic purposes. However, we note that an agent may issue multiple connect calls in the environment, and the frequency of such calls is application dependent.

\subsection{Defining The Response Determinant Function}\label{subsec:resp_det_func}
Notice that using $\vartheta$ ensures that $ i$ considers only a single insertion point during the connection process. Nonetheless, since $j$ could also be broadcasting Announce messages, there is a need for a mechanism that avoids cyclic connections in the hierarchy. To this end, the result of $\phi_i: A\rightarrow \mathbb{B}$ must be consistent and independent of the problem horizon. We define an instance of $\phi$ in equation~\ref{eq:phi} based on the ordered set of agents mentioned in Section~\ref{subsec:adhoc-def}.

\begin{equation}
    \phi_i(j)=
    \left\{ \begin{array}{ll}
        \mathsf{TRUE}       & \quad \text{if } i < j, \\
        \mathsf{FALSE}  & \quad \text{Otherwise }
    \end{array} \right.
    \label{eq:phi}
\end{equation}

This definition has the intuition that the agent with the lowest index will be the root. In the generation of the multi-agent hierarchy, two extreme cases must be considered. The first is a situation where the hierarchy tends towards a chain. Such a scenario may be helpful for synchronized computations but undesirable when parallelism is desired. The second is when all agents are connected to one root (tree of depth 1). While this encourages parallelism, removing the root causes all other agents to be affected and the need to invoke the connection algorithm. To balance these extremes, we use a max-degree heuristic to limit the possible number of neighbors and define $\vartheta$ to weight agents with a single or no child higher when selecting from the response list.

\section{Experiments}\label{sec:experiments}
\subsection{Setting}
We used C-CoCoA~\cite{SarkerAmit;ChoudhuryMoumita;khan2021} and applied similar non-linear optimization methods in~\cite{SarkerAmit;ChoudhuryMoumita;khan2021} to the SDPOP (C-SDPOP) to obtain representative top-down and bottom-up DCOP algorithms, respectively, in our experiments\footnote{\url{https://github.com/bbrighttaer/ddcop-dynagraph}}. We defined a binary constraint between two agents as $f_k(x,y)= ax^2+bxy+cy^2$ where $x$ is the value of $i$ and $y$ is the value of $j$. The coefficients were sampled randomly between the range $[-5, 5]$. The lower and upper bounds for each agent's domain were set at $[-50, 50]$ and each agent samples three values from this range to constitute its domain. The change-constraint event randomly selects a constraint between two agents and changes the coefficients of the constraint $(a, b, c)$ by sampling randomly from the range $[-5, 5]$.

We restricted the maximum number of children an agent could have in the multi-agent hierarchy (max-out degree) to 3, 5, and 6 to create different experiment settings for the evaluation. For each max-out degree, we generated five random problems where the first 100 events add an agent into the environment, and the subsequent 10 events were change-constraint events. Then 10 agent removal events were executed in the environment. We applied a delay of 5 seconds in between events to allow time for an applied event's effect to propagate before gathering the respective metrics. The metrics we considered in our experiments are solution cost and the total number of messages. Each generated problem was ran with 10 different random seeds. The reported results are averages across the different settings of a given max-out degree.

\begin{figure}[ht!]
     \centering
     \begin{subfigure}[]{0.48\textwidth}
         \centering
         \includegraphics[width=\textwidth]{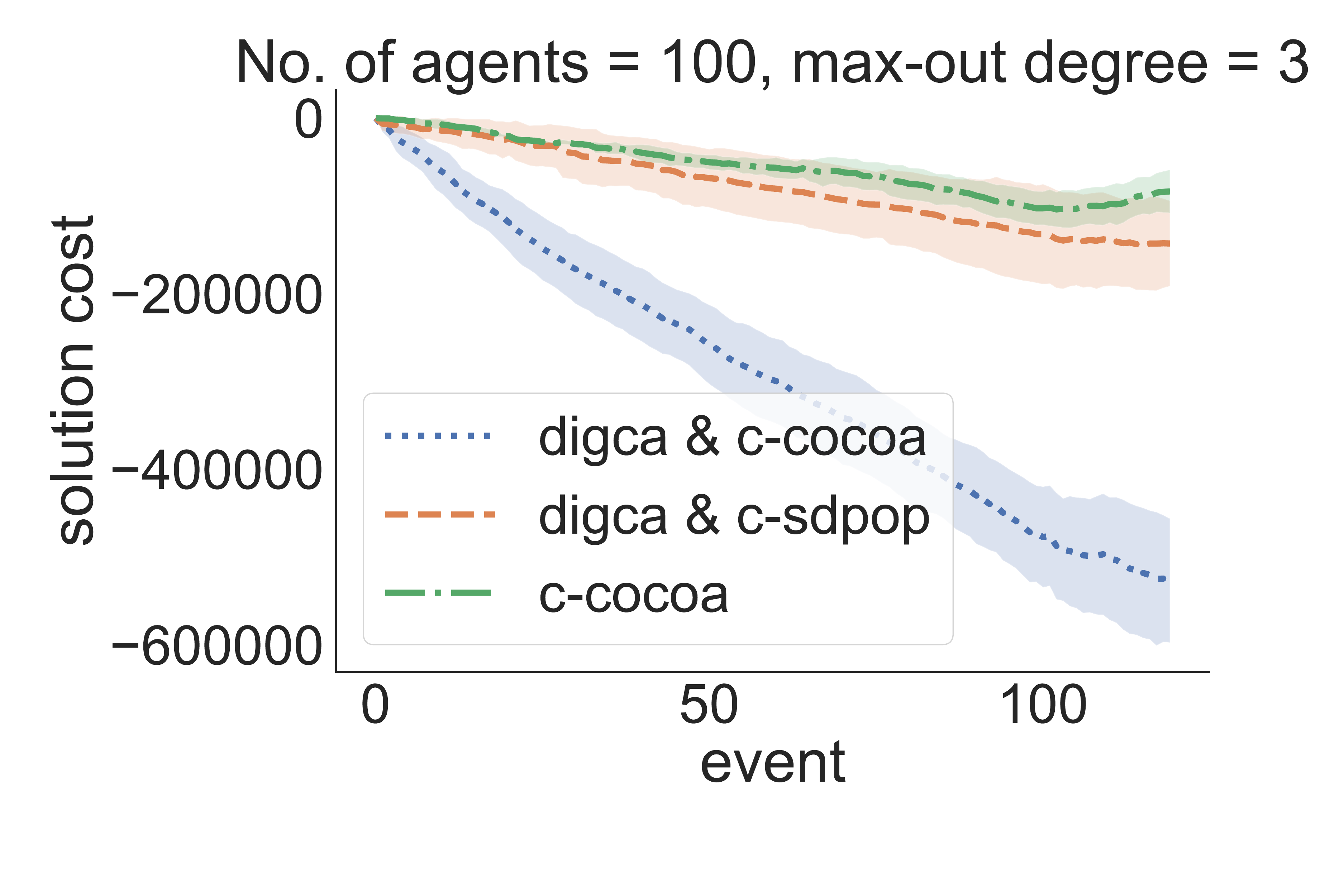}
         \label{fig:cost_d5}
     \end{subfigure}
     
     \begin{subfigure}[]{0.48\textwidth}
         \centering
         \includegraphics[width=\textwidth]{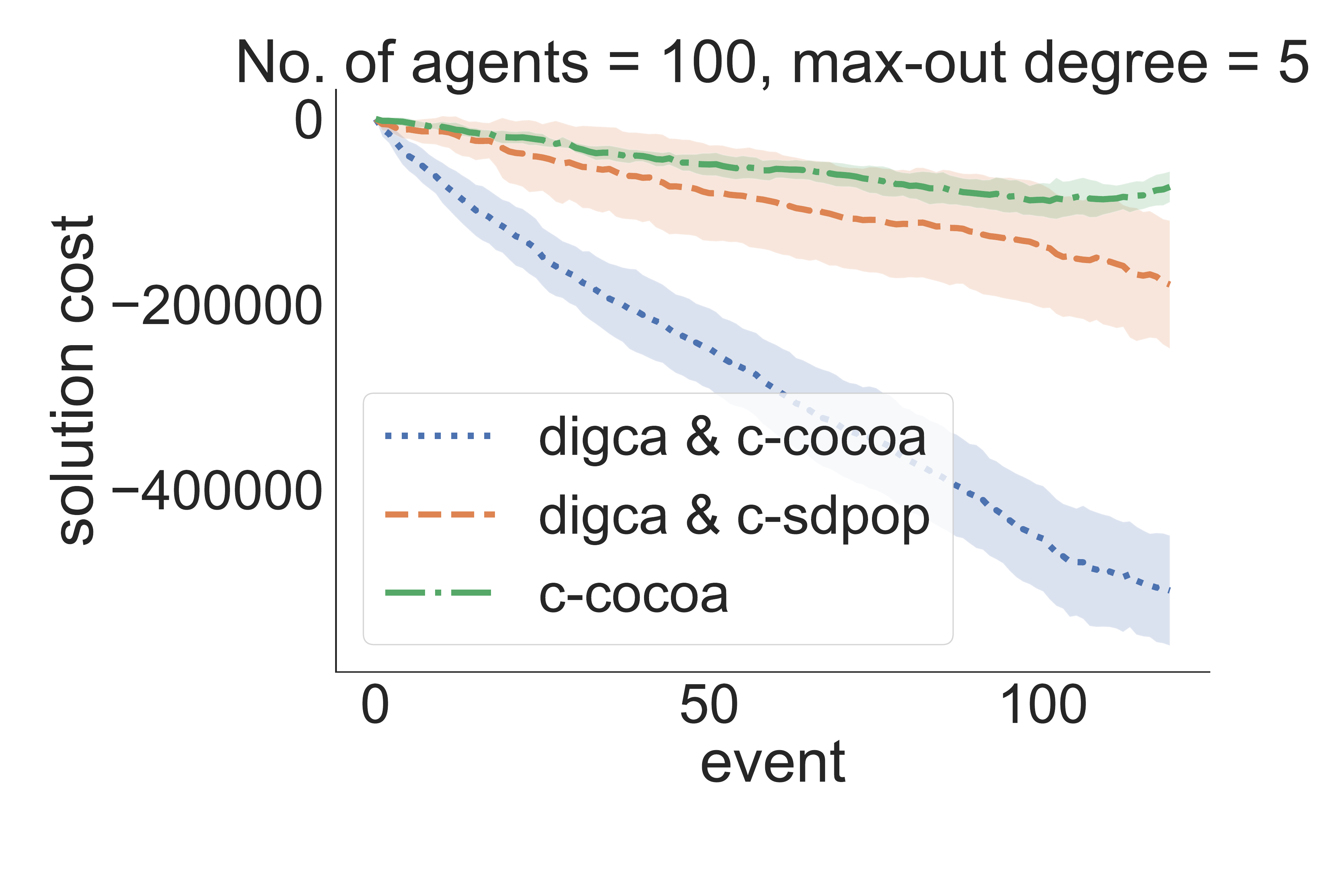}
         \label{fig:cost_d5}
     \end{subfigure}
     \begin{subfigure}[]{0.48\textwidth}
         \centering
         \includegraphics[width=\textwidth]{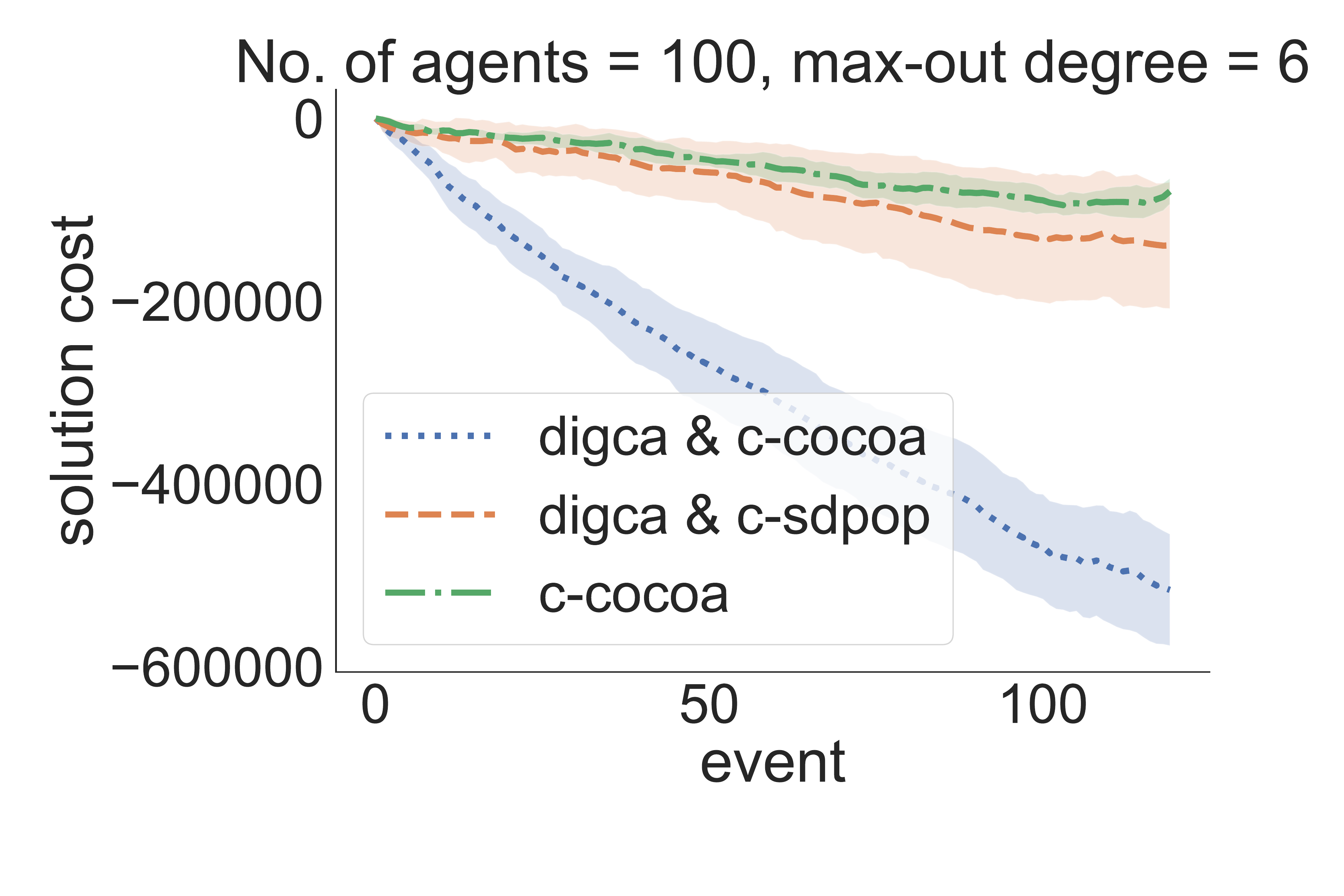}
         \label{fig:cost_d6}
     \end{subfigure}
     
     \caption{Solution cost result after 120 events in the environment. The first 100 events are add-agent events, the next 10 events are change-constraint events, and the remaining 10 events are remove-agent events.}
     \label{fig:experiment_results}
\end{figure}

\subsection{Results}
The solution cost results for the three experiment settings are shown in Figure~\ref{fig:experiment_results}. Also, the average number of messages resulting from the interactions over the horizon are shown in Table~\ref{tab:total-num-messages}. The standard deviations are shown in parenthesis. Generally, the solution cost performance of the DIGCA was stable in the different experiment settings considered. In contrast, the C-CoCoA algorithm performed poorly in all the experiments. On the solution cost, we observed that pairing our proposed approach with the C-CoCoA reduced the cost in all three experiments.

Similarly, we observed that C-CoCoA used almost twice as many messages during the simulation compared to CoCoA-DIGCA in all experiment settings. This observation is not surprising since the C-CoCoA algorithm is a classical DCOP algorithm and therefore restarts for every event. Thus, this performance shows the importance of having a well-defined structure for agent execution in an open and dynamic environment as enabled by DIGCA. In the case of C-SDPOP, while the solution cost was not as low, it used the fewest messages in all experiments. Aside from interaction complexity reduction enabled by DIGCA, the C-SDPOP algorithm also uses a fault-containment procedure~\cite{Petcu2005} to reduce the number of messages forwarded to the root agent. We realized in our experiments that this fault-containment procedure was necessary to avoid the root agent triggering the propagation of VALUE messages down the entire hierarchy once a UTIL message percolates up. Thus, for algorithms that use multiple rounds of message passing in the hierarchy for the optimization process, there is a need for local agent mechanisms to determine which sub-trees receive change propagation. We also note that the usage of DIGCA with different DCOP algorithms highlights its effectiveness in facilitating the development of multi-agent coordination approaches for an open and dynamic environment.

\begin{table}[!ht]
    \centering
    \caption{Average number of messages by algorithms with standard deviations}
    \label{tab:total-num-messages}
    \begin{tabular}{|l|lll|}
    \hline
    \multicolumn{1}{|c|}{\multirow{2}{*}{Algorithm}} & \multicolumn{3}{c|}{Maximum out degree}                                                                    \\ \cline{2-4} 
    \multicolumn{1}{|c|}{}                           & \multicolumn{1}{c|}{3}                  & \multicolumn{1}{c|}{5}                  & \multicolumn{1}{c|}{6} \\ \hline
    DIGCA and C-CoCoA                                & \multicolumn{1}{l|}{4435.958 (273.067)} & \multicolumn{1}{l|}{4596.480 (412.163)} & 4595.617 (324.378)     \\ \hline
    DIGCA and C-SDPOP                                & \multicolumn{1}{l|}{4301.940 (407.647)} & \multicolumn{1}{l|}{4265.408 (392.879)} & 4108.081 (361.208)     \\ \hline
    C-CoCoA                                          & \multicolumn{1}{l|}{74576 (2099.177)}   & \multicolumn{1}{l|}{76066 (3682.198)}   & 74920 (3281.262)       \\ \hline
    \end{tabular}
\end{table}

\section{Conclusion}\label{sec:conclusion}
In this paper, we have discussed the ad-hoc distributed multi-agent hierarchy generation problem. We have also proposed a distributed algorithm for constructing and maintaining a stable multi-agent hierarchy for interaction when collaborating in a dynamic environment. Our proposed approach addresses a vital issue in multi-agent operations in open and dynamic environments. Unlike existing methods, DIGCA does not require an existing interaction graph or reconstruction of the entire multi-agent hierarchy when changes are detected. The method's effectiveness has been shown using variants of well-known DCOP proposals. Our future work would apply the proposed method to problems in DCOP\_MSTs and other practical domains.

\section*{Acknowledgments}
We thank David Livingstone Hini for designing the web interface for the experiments.

\bibliographystyle{unsrt}  
\bibliography{references}

\begin{thebibliography}{10}

\bibitem{Yedidsion2018}
Harel Yedidsion, Roie Zivan, and Alessandro Farinelli.
\newblock {Applying max-sum to teams of mobile sensing agents}.
\newblock {\em Engineering Applications of Artificial Intelligence}, 71:87--99,
  may 2018.

\bibitem{973385}
P.E. Rybski, S.A. Stoeter, M.~Gini, D.F. Hougen, and N.~Papanikolopoulos.
\newblock Effects of limited bandwidth communications channels on the control
  of multiple robots.
\newblock In {\em Proceedings 2001 IEEE/RSJ International Conference on
  Intelligent Robots and Systems. Expanding the Societal Role of Robotics in
  the the Next Millennium (Cat. No.01CH37180)}, volume~1, pages 369--374 vol.1,
  2001.

\bibitem{10.1145/1160633.1160885}
Paritosh Padhy, Rajdeep~K. Dash, Kirk Martinez, and Nicholas~R. Jennings.
\newblock A utility-based sensing and communication model for a glacial sensor
  network.
\newblock In {\em Proceedings of the Fifth International Joint Conference on
  Autonomous Agents and Multiagent Systems}, AAMAS '06, page 1353–1360, New
  York, NY, USA, 2006. Association for Computing Machinery.

\bibitem{10.1007/978-3-642-40643-0_21}
Marc Pujol-Gonzalez, Jes{\'u}s Cerquides, Pedro Meseguer, Juan~Antonio
  Rodr{\'i}guez-Aguilar, and Milind Tambe.
\newblock Engineering the decentralized coordination of uavs with limited
  communication range.
\newblock In Concha Bielza, Antonio Salmer{\'o}n, Amparo Alonso-Betanzos,
  J.~Ignacio Hidalgo, Luis Mart{\'i}nez, Alicia Troncoso, Emilio Corchado, and
  Juan~M. Corchado, editors, {\em Advances in Artificial Intelligence}, pages
  199--208, Berlin, Heidelberg, 2013. Springer Berlin Heidelberg.

\bibitem{10.5555/1402298.1402308}
Robert Junges and Ana L.~C. Bazzan.
\newblock Evaluating the performance of dcop algorithms in a real world,
  dynamic problem.
\newblock In {\em Proceedings of the 7th International Joint Conference on
  Autonomous Agents and Multiagent Systems - Volume 2}, AAMAS '08, page
  599–606, Richland, SC, 2008. International Foundation for Autonomous Agents
  and Multiagent Systems.

\bibitem{Nguyen2012}
Trung~Thanh Nguyen and Xin Yao.
\newblock {Continuous dynamic constrained optimization-the challenges}.
\newblock {\em IEEE Transactions on Evolutionary Computation}, 16(6):769--786,
  2012.

\bibitem{Sultanik2009}
Evan~A. Sultanik, Robert~N. Lass, and William~C. Regli.
\newblock {Dynamic configuration of agent organizations}.
\newblock {\em IJCAI International Joint Conference on Artificial
  Intelligence}, pages 305--311, 2009.

\bibitem{Yeoh2015}
William Yeoh, Pradeep Varakantham, Xiaoxun Sun, and Sven Koenig.
\newblock {Incremental DCOP search algorithms for solving dynamic DCOP
  problems}.
\newblock {\em Proceedings - 2015 IEEE/WIC/ACM International Joint Conference
  on Web Intelligence and Intelligent Agent Technology, WI-IAT 2015},
  2:257--264, 2015.

\bibitem{Petcu2005}
Adrian Petcu and Boi Faltings.
\newblock {Superstabilizing, fault-containing distributed combinatorial
  optimization}.
\newblock {\em Proceedings of the National Conference on Artificial
  Intelligence}, 1:449--454, 2005.

\bibitem{Yeoh2011}
William Yeoh, Pradeep Varakantham, Xiaoxun Sun, and Sven Koenig.
\newblock {Incremental DCOP Search Algorithms for Solving Dynamic DCOPs (
  Extended Abstract )}.
\newblock {\em Artificial Intelligence}, (Aamas):1069--1070, 2011.

\bibitem{Barambones2021}
Jose Barambones, Ricardo Imbert, and Cristian Moral.
\newblock {Applicability of multi-agent systems and constrained reasoning for
  sensor-based distributed scenarios: A systematic mapping study on dynamic
  DCOPs}.
\newblock {\em Sensors}, 21(11), 2021.

\bibitem{Zivan2015}
Roie Zivan, Harel Yedidsion, Steven Okamoto, Robin Glinton, and Katia Sycara.
\newblock {Distributed constraint optimization for teams of mobile sensing
  agents}.
\newblock {\em Autonomous Agents and Multi-Agent Systems}, 29(3):495--536,
  2015.

\bibitem{YoussefHamadiChristianBessiere1998}
Joel~Quinqueton {Youssef Hamadi Christian Bessiere}.
\newblock {Backtracking in Distributed Constraint Networks}.
\newblock {\em ECAI}, pages 219--223, 1998.

\bibitem{COLLIN1994297}
Zeev Collin and Shlomi Dolev.
\newblock {Self-stabilizing depth-first search}.
\newblock {\em Information Processing Letters}, 49(6):297--301, 1994.

\bibitem{Petcu2007}
Adrian Petcu and Boi Faltings.
\newblock {Optimal solution stability in dynamic, distributed constraint
  optimization}.
\newblock {\em Proceedings of the IEEE/WIC/ACM International Conference on
  Intelligent Agent Technology, IAT 2007}, pages 321--327, 2007.

\bibitem{Fioretto2018}
Ferdinando Fioretto, Enrico Pontelli, and William Yeoh.
\newblock {Distributed constraint optimization problems and applications: A
  survey}.
\newblock {\em Journal of Artificial Intelligence Research}, 61:623--698, 2018.

\bibitem{SarkerAmit;ChoudhuryMoumita;khan2021}
Md~Mosaddek {Sarker, Amit; Choudhury, Moumita; khan}.
\newblock {A Local Search Based Approach to Solve Continuous DCOPs}.
\newblock {\em Aamas 21}, pages 1127--1135, 2021.

\end{thebibliography}

\end{document}